\documentclass{article}
\pdfoutput=1

\usepackage[preprint]{neurips_2024}

\usepackage[colorlinks=true, linkcolor=blue, citecolor=blue, urlcolor=blue]{hyperref}

\usepackage[utf8]{inputenc} 
\usepackage[T1]{fontenc}    
\usepackage{hyperref}       
\usepackage{url}            
\usepackage{booktabs}       
\usepackage{amsfonts}       
\usepackage{ulem}           
\usepackage{nicefrac}       
\usepackage{microtype}      
\usepackage{xcolor}         
\usepackage{amsmath}
\usepackage{float}
\usepackage{tabularray}
\usepackage{multirow}
\usepackage{booktabs, caption, array}
\usepackage{amssymb}
\usepackage{lipsum}         
\usepackage{CJKutf8}        
\usepackage{nomencl}
\usepackage{glossaries}
\usepackage{xspace}
\usepackage{multirow}
\usepackage{graphicx}
\usepackage{bbding}
\usepackage{pifont}
\usepackage{algorithm}
\usepackage{subcaption}
\usepackage{enumitem}
\usepackage{multicol}
\usepackage{caption}
\usepackage{array}
\usepackage{fp}
\usepackage{multirow}   
\usepackage{makecell}
\usepackage{flushend}
\usepackage{cases}
\usepackage{color}
\usepackage{algpseudocode}
\usepackage{listings}
\usepackage{mathrsfs}
\usepackage{longtable}
\usepackage{colortbl}
\usepackage{bm}
\usepackage{caption}

\usepackage{varwidth} 
\setlist[itemize]{noitemsep, topsep=0pt}
\usepackage{arydshln}
\usepackage{lipsum}
\definecolor{codegreen}{rgb}{0,0.3,0.6}
\definecolor{codegray}{rgb}{0.5,0.5,0.5}

\usepackage{dashrule}

\usepackage[dvipsnames,svgnames,x11names,table]{xcolor}
\usepackage[most]{tcolorbox}
\definecolor{codegreen}{rgb}{0,0.3,0.6}
\definecolor{codegray}{rgb}{0.5,0.5,0.5}

\newcommand{\ignore}[1]{}

\definecolor{darkorange}{RGB}{255, 140, 0}

\usepackage{relsize}
\usepackage{stfloats}
\usepackage{tabularx}

\definecolor{myorange}{RGB}{255,127,0}
\newtcolorbox{promptbox}[3][Judge Prompt]{
colback=black!5!white,
arc=5pt, 
boxrule=0.5pt,
fonttitle=\bfseries,
title=#1, 
before upper={\small}, fontupper=\fontfamily{ptm}\selectfont,
colframe=#2,
label=#3,
}

\title{SELF-EMO: Emotional Self-Evolution from Recognition to Consistent Expression}

\author{%
  Shaowei Zhang$^{1}$\thanks{Work done during internship at SenseTime}~,
  Faqiang Qian$^{1}$\thanks{Project leader},
  Yan Chen$^{1,2}$\footnotemark[1]~,
  Ziliang Wang$^{1}$,\\
  \textbf{Kang An}$^{1,3}$\footnotemark[1]~,
  \textbf{Yong Dai}$^{4}$,
  \textbf{Mengya Gao}$^{1}$,
  \textbf{Yichao Wu}$^{1}$
  \\
  $^1$SenseTime \\
  $^2$Tsinghua University $^3$Shanghai Jiao Tong University $^4$X-Humanoid \\
  \texttt{\{qianfaqiang,gaomengya,wuyichao\}@senseauto.com}\\
}

\begin{document}

\maketitle

\begin{abstract}
Emotion Recognition in Conversation (ERC) has become a fundamental capability for large language models (LLMs) in human-centric interaction. Beyond accurate recognition, recent studies highlight the importance of coherent emotional expression, yet both abilities are severely limited by the scarcity and static nature of high-quality annotated data.In this work, we propose \textbf{SELF-EMO}, a self-evolution framework grounded in a psychologically motivated hypothesis: \texttt{the better a model predicts others’ emotions, the better it can generate its own emotionally consistent responses}. Building on this insight, we explicitly incorporate two auxiliary tasks—\texttt{emotional understanding} and \texttt{emotional expression}—and formulate a role-based self-play paradigm in which the model simultaneously acts as an emotion recognizer and a dialogue responder. Through iterative interaction, the agents continuously generate diverse conversational trajectories, forming a scalable data generation process. To ensure training quality, we construct a data flywheel mechanism: each self-play rollout produces multiple candidate emotional predictions and responses, which are then filtered via a smoothed IOU-based reward to select high-quality samples. The selected samples are fed back into training, enabling continuous self-improvement without external supervision. Based on this paradigm, we further develop \textbf{SELF-GRPO}, a reinforcement learning algorithm that stabilizes optimization under diverse emotional outputs by combining multi-label alignment rewards with group-level consistency signals.Extensive experiments on three benchmark datasets (IEMOCAP, MELD, and EmoryNLP) demonstrate that SELF-EMO achieves state-of-the-art (SOTA) performance without relying on external retrieval or auxiliary models. Under a unified training setting, our method improves average accuracy by \textbf{+6.33\%} on Qwen3-4B and \textbf{+8.54\%} on Qwen3-8B, with consistent gains across all benchmarks. These results validate the effectiveness of the proposed self-evolution paradigm and highlight its strong generalization ability.
\end{abstract}

\section{Introduction}
Large language models (LLMs) are increasingly embedded in daily human life—not merely as tools, but as conversational partners~\cite{gpt4,qwen3,llama3,deepseekv3}. This shift makes emotionally intelligent interaction a pressing research priority, with Emotion Recognition in Conversation (ERC) emerging as a core task for building more anthropomorphic AI systems~\cite{ercmodeling}.

\begin{figure}[h]
    \centering
    \includegraphics[width=0.9\linewidth]{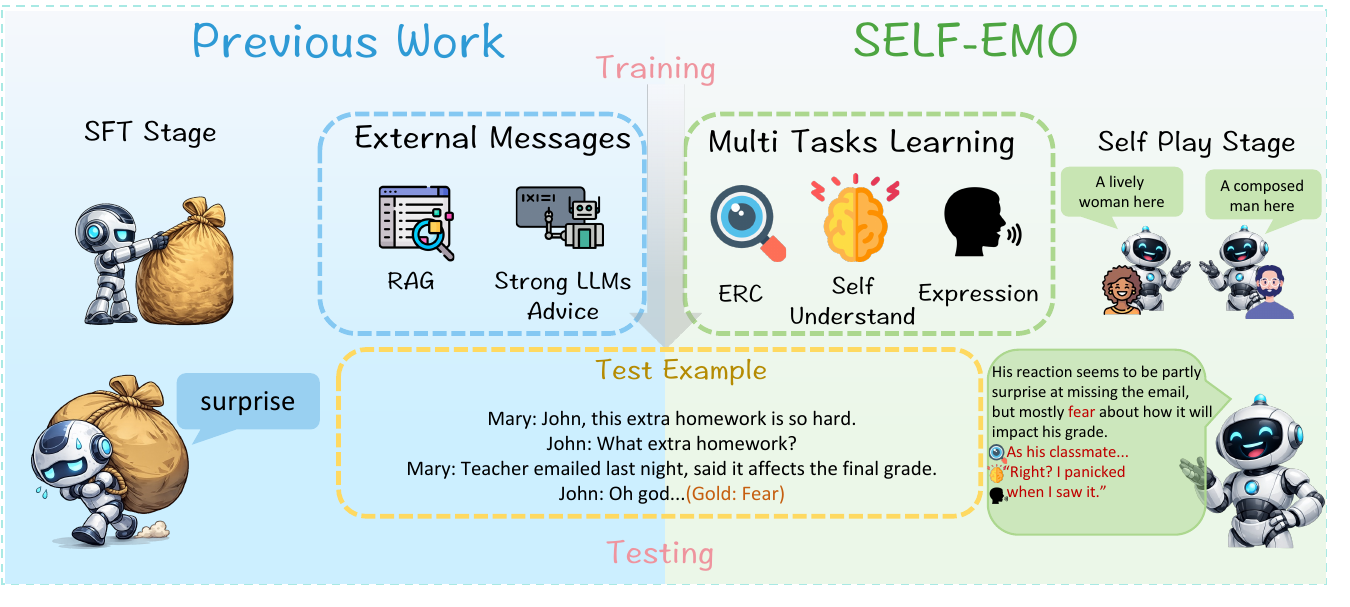}
    \caption{Comparison of prior work and \textbf{SELF-EMO}, a multi-task self-evolution framework that unifies emotion recognition, emotional understanding, and emotion-consistent response generation within a closed-loop self-play paradigm.}
    \label{fig:introduce}
\end{figure}

Despite recent progress, existing LLM-based ERC methods remain fundamentally limited by their reliance on static external knowledge. Approaches based on retrieval-augmented generation (RAG) or auxiliary LLM supervision ~\cite{lei2024instructercreformingemotionrecognition,li2025llmsfeelteachingemotion,Xue_2024} improve in-distribution performance, but introduce a strong dependency on fixed datasets and additional models. As a result, these methods fail to generalize across datasets, with performance dropping significantly under cross-domain settings. More importantly, these limitations point to a deeper bottleneck: \texttt{the lack of a scalable training signal}. Current ERC pipelines rely on static annotated datasets, which cannot provide the diverse and continuously evolving supervision required for iterative optimization. This creates a fundamental tension between performance, efficiency, and generalization, preventing models from achieving robust emotional intelligence.

To address this challenge, we propose \textbf{SELF-EMO}, a self-play reinforcement learning framework that transforms emotional understanding into a scalable training signal. Our approach is grounded in a psychologically motivated hypothesis that is theoretically justified and empirically validated in this work: \texttt{the better a model predicts others’ emotions, the better it can generate its own emotionally consistent responses}.Building on this insight, we extend the standard ERC formulation by introducing two auxiliary tasks—\texttt{emotional understanding} and \texttt{emotional expression}—forming a structured pipeline from recognition to response generation. We organize these tasks into a \textbf{role-based self-play paradigm}, where persona-conditioned agents act as both an emotion recognizer and a dialogue responder. Through iterative interaction, the model generates diverse conversational trajectories with emotion predictions, self-emotional states, and responses.To enable scalable self-improvement, we further construct a \textbf{data flywheel mechanism}. In each self-play rollout, multiple candidate predictions and responses are generated, evaluated via a \textbf{smoothed IOU-based reward}, and high-quality samples are selected and fed back into training. This forms a closed loop of \texttt{generation $\rightarrow$ selection $\rightarrow$ reuse}, allowing continuous refinement without external supervision.To optimize this process under diverse emotional outputs, we introduce \textbf{SELF-GRPO}, a multi-emotion reinforcement learning algorithm that combines multi-label alignment rewards with group-level consistency signals for more stable credit assignment. This design provides both theoretical and empirical support for the proposed self-evolution paradigm.Our main contributions are summarized as follows:
\begin{itemize}
    \item \textbf{SELF-EMO Framework:} We propose a multi-task self-evolution framework that unifies emotion recognition, emotional understanding, and emotion-consistent response generation within a role-based self-play paradigm, enabled by a generate--filter--reuse data flywheel for continuous self-improvement.
    
    \item \textbf{SELF-GRPO Algorithm:} We develop a multi-emotion reinforcement learning algorithm that enhances credit assignment via core emotion set aggregation and group-level consistency modeling, improving stability under diverse predictions.
    
    \item \textbf{Empirical Performance and Generalization:} Extensive experiments show that our approach achieves state-of-the-art performance with strong cross-dataset generalization, yielding accuracy gains of \textbf{+6.33\%} on Qwen3-4B and \textbf{+8.54\%} on Qwen3-8B.
\end{itemize}

\section{Related Work}
\subsection{ERC Methods}
Prior work in the field of ERC has primarily concentrated on three methodological paradigms: Recurrent Neural Networks (RNNs)~\cite{rnn}, Graph Neural Networks (GNNs)~\cite{gnn}, and Pre-trained Language Models (PLMs)~\cite{bert}.
In RNN-based studies, DialogueRNN~\cite{diarnn} and DialogueCRN~\cite{dialoguecrn} adopt RNN architectures to track contextual information and the individual states of speakers throughout the conversation. In GNN-based studies, models capture relational information by constructing graph structures among utterances and speakers. For example, DialogueGCN~\cite{ghosal2019dialoguegcngraphconvolutionalneural} treats utterances as graph nodes and builds edges based on contextual information, while DAG-ERC~\cite{shen2021directedacyclicgraphnetwork} further incorporates speaker identity and utterance position into the graph. In studies based on pre-trained language models, BERT-ERC~\cite{berterc} leverages models such as BERT and improves performance by introducing a fine-grained emotion classification module and a two-stage training strategy.

With the rise of LLMs, current research in the ERC field mostly uses LLMs combined with retrieval-augmented methods to improve model performance on the ERC task, among which InstructERC~\cite{lei2024instructercreformingemotionrecognition} is the first study to introduce these methods in this field. Subsequently, BiosERC~\cite{Xue_2024} further adds character traits to the input and adopts LoRA~\cite{lora} training to efficiently fine-tune the model. Furthermore, PRC-Emo~\cite{li2025llmsfeelteachingemotion} proposes to additionally include internal and external emotional descriptions of speakers in the input and employs a curriculum learning strategy to enhance model performance. However, these works introduce static knowledge bases for retrieval and incorporate external knowledge analysis, which also lead to problems of poor model generalization ability and high inference resource consumption.

\subsection{Self-Play for Self-Evolution}
Due to the rise of LLM-based methods, an increasing number of works have begun to focus on how to achieve a closed training loop solely through data generated by the model itself under the constraint of limited high-quality human-crafted data~\cite{selfplay1,selfplay2,selfplay3}, thereby significantly reducing the cost of manual data creation. This approach has yielded substantial benefits in multiple domains. For instance, AbsoluteZero~\cite{absolute} was proposed in the code generation domain, training the model itself as a user instruction generator to obtain high-quality problems while simultaneously training its ability to solve the generated problems. In the mathematical reasoning domain, works such as SVS~\cite{svs} and PromptCot~\cite{promptcot} have also demonstrated that the model itself can serve as a high-quality problem generator, producing problems of appropriate difficulty for the model to perform reinforcement learning and improve problem-solving performance. However, self-play works all need to address a fundamental issue: how to design a logically rigorous operational paradigm for self-play to achieve a closed training loop.

\subsection{Psychological Connections of Emotion Tasks}
At present, there have already been a large number of psychological studies that have made detailed arguments and experiments on the connection among the three tasks of emotion understanding and recognition, self-emotion generation, and expression. Among them, the theory of levels of emotional awareness ~\cite{lane1987levels} indicates that individuals must first recognize, differentiate, and label emotions before they can conduct more complex emotional reasoning and social emotion regulation ~\cite{fonagy2018affect}, and the differentiation of others’ emotions is the core of explaining others’ behaviors and generating appropriate responses. On the contrary, existing clinical evidence ~\cite{psychological_evidence} shows that the lack of emotional differentiation ability will directly impair empathy and expression. These studies all indicate that emotion understanding and recognition is the premise of self-emotion generation and expression. Moreover, studies ~\cite{van2015social} show that others’ emotions will influence the observer’s emotions and behavioral decisions, and the occurrence of this phenomenon can be summarized as that how one understands others’ emotions will directly control self-emotion generation and expression.

\section{Methodology}
\begin{figure}[t]
    \centering
    \includegraphics[width=1.0\linewidth]{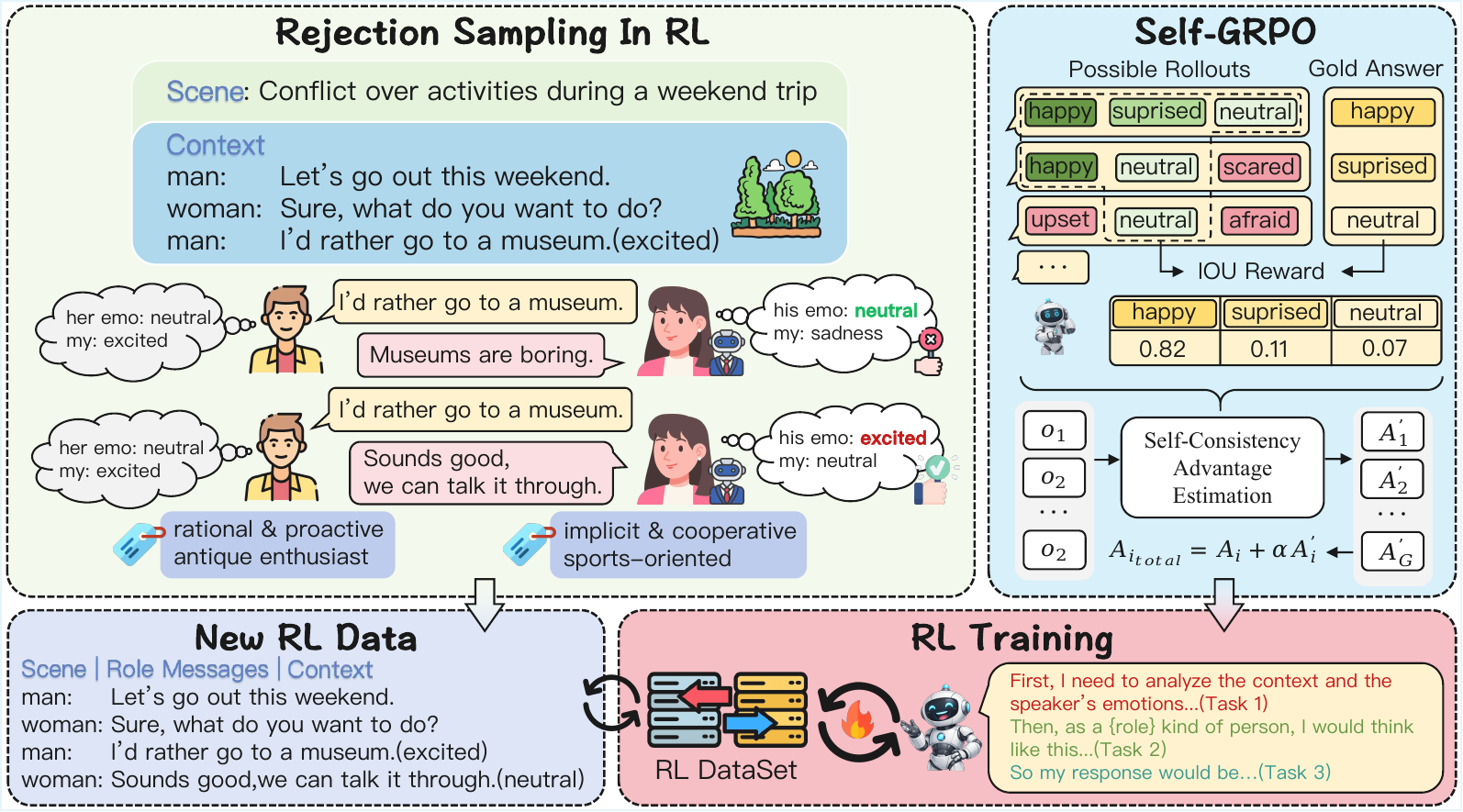}
    \caption{The overall framework of SELF-EMO.During SELF-GRPO tuning, the model explicitly optimizes the ERC task and implicitly optimizes the other two. High-quality samples generated during training are retained for subsequent training.}
    \label{fig:method}
\end{figure}

In this section, we detail the SELF-EMO training framework, which extracts rich situational and personality information from existing datasets and constructs a closed-loop evolution paradigm across three key psychological tasks: Emotion Recognition in Conversation (ERC), emotional understanding, and emotional expression. We first provide the theoretical proof for SELF-EMO’s semi-supervised mechanism. Subsequently, we describe the algorithmic process of SELF-EMO and the SELF-GRPO reinforcement learning algorithm designed to support self-play data generation.

\subsection{Semi-supervised Multi-task Entropy Reduction}
\label{subsec:theory}

We introduce two unsupervised auxiliary tasks to support the ERC task $T_o$: \texttt{self-personality understanding} ($T_s$) and \texttt{self-emotion expression} ($T_r$). Given a dialogue context $C$ and a personality trait $PI$, the model produces three outputs sequentially:
\begin{equation}
T_{o} = \pi_{\theta}(e_{o} \mid C, PI), \quad
T_{s} = \pi_{\theta}(e_{s} \mid C, PI, e_{o}), \quad
T_r = \pi_{\theta}(r \mid C, PI, e_{o}, e_{s})
\end{equation}
where $e_o$ is the predicted emotion of the interlocutor, $e_s$ is the model's self-perceived emotional response, and $r$ is the self-emotion expression generated by $T_r$.

\paragraph{Theoretical Derivation}
Drawing on cognitive appraisal theory, an agent's internal state is shaped by its appraisal of others' emotions. We model this as:
\begin{equation}
T_{s} = f_{s}(T_{o}, C, PI) + \epsilon_{s}, \qquad T_r = f_{r}(T_{o}, T_{s}, C, PI) + \epsilon_{r}
\end{equation}
where $f_s$ and $f_r$ are the appraisal and expression mappings, and $\epsilon_s$, $\epsilon_r$ are residual terms.

From an information-theoretic perspective~\cite{oneentropy}, optimizing $T_o$ via reinforcement learning reduces the output entropy $H_{\pi_\theta}(e_o \mid C, PI)$. Given the functional dependencies above, this propagates to the auxiliary tasks:
\begin{equation}
\downarrow H_{\pi_{\theta}}(e_{o} \mid C, PI)
\;\Rightarrow\;
\downarrow H_{\pi_{\theta}}(e_{s} \mid C, PI, e_{o})
\;\Rightarrow\;
\downarrow H_{\pi_{\theta}}(r \mid C, PI, e_{o}, e_{s})
\end{equation}
This shows that supervised training on $T_o$ implicitly reduces uncertainty in $T_s$ and $T_r$, providing a theoretical basis for the unsupervised components in SELF-EMO.

\subsection{Extracting Personality Information and Constructing Dialogue Data}

Let $\mathcal{D} = \{\mathcal{D}_{\text{IEMOCAP}}, \mathcal{D}_{\text{MELD}}, \mathcal{D}_{\text{EmoryNLP}}\}$ be the collection of source datasets, and let $d \in \mathcal{D}$ denote a sampled dialogue segment. We use a large language model $M_e$, guided by prompt $P_{\text{extrac}}$, to extract speaker personality information $PI$ from each segment:
\begin{equation}
PI = M_e(d, P_{\text{extrac}}).
\end{equation}
To improve coverage and diversity, $M_e$ is instantiated with two models: Seed-1.8, DeepSeek-V3.2.

Given historical context $\mathit{C}$ and personality information $PI$, a generation model $\pi_\theta$ produces a structured dialogue sample $R^{\text{syn}}$ through three sequential steps:
\begin{equation}
R^{\text{syn}} = \left[
  \underbrace{e_o = \pi_\theta(\mathit{C},\, PI)}_{\text{Others' Emotion Recognition}},\;
  \underbrace{e_s = \pi_\theta(\mathit{C},\, PI,\, e_o)}_{\text{Self-Emotion Generation}},\;
  \underbrace{r = \pi_\theta(\mathit{C},\, PI,\, e_o,\, e_s)}_{\text{Self-Emotion Expression}}
\right],
\end{equation}
where $e_{\text{o}}$ is the recognized emotion of the interlocutor and $e_{\text{s}}$ is the speaker's generated self-emotion, and $r$ is the final response. Since the three steps produce non-overlapping outputs, $\pi_\theta$ generates all content in a single pass under prompt $P_{\text{gen}}$. The resulting samples $R^{\text{syn}}$ serve as initialization data for the self-play training stage. All prompts used above can be found in Appendix~\ref{appendix:prompts}.

\subsection{Data Closed-loop Self-play Reinforcement Learning Fine-tuning}

During online sampling, for each input $(PI, C_{ori})$, the model produces:
\begin{equation}
\begin{aligned}
(e_o,e_s,r,Output_{Dict}) &\sim \pi_\theta(e_o,e_s,r \mid C_{ori}, PI) \\
Prediction_o,\, Label_s,\, r &\sim Extract(Output_{Dict})
\end{aligned}
\end{equation}
where $e_o$ contains the emotion prediction $Prediction_o$ for the last speaker, $e_s$ contains the model's self-emotion label $Label_s$ under the given personality, and $r$ is the generated response. High-quality samples are used to construct new training data:
\begin{equation}
\label{eq:update}
C_{new} \leftarrow C_{ori} \cup r, \quad Label_{new} \leftarrow Label_{s}
\end{equation}

To handle the discrete ERC label space, we propose a \textbf{multi-label weighted reward}. The model outputs up to $N\!\le\!3$ emotion predictions, each with a weight $Weight^{emo_n}$, yielding a smoothed IOU-based reward:
\begin{equation}
\label{eq:reward}
\begin{aligned}
Prediction_o &= \{(Prediction^{emo_n}, Weight^{emo_n})\}_{n=1}^{N} \\
\mathcal{R}(\cdot) &=
\begin{cases}
0, & rule(Output_{Dict}) = 0 \\
IOU(Prediction_o,\, Label_{ori|new}) + 0.1, & rule(Output_{Dict}) = 1
\end{cases}
\end{aligned}
\end{equation}
Let $P = \{(e,\, w^P_e)\}$ and $L = \{(e,\, w^L_e)\}$ denote the raw prediction and label weight sets. We normalize each to sum to 1:
\begin{equation}
\hat{w}^P_e = \frac{w^P_e}{\sum_{e'} w^P_{e'}}, \quad \hat{w}^L_e = \frac{w^L_e}{\sum_{e'} w^L_{e'}}
\end{equation}
Setting $\hat{w}^P_e = 0$ for $e \notin P$ and $\hat{w}^L_e = 0$ for $e \notin L$, the weighted IOU is:
\begin{equation}
\text{IOU}(P, L) = \frac{\sum_{e \in P \cup L} \min(\hat{w}^P_e,\, \hat{w}^L_e)}{\sum_{e \in P \cup L} \max(\hat{w}^P_e,\, \hat{w}^L_e)}
\end{equation}
This gives partial credit for near-correct predictions.
\paragraph{SELF-GRPO}
We extend GRPO~\cite{grpo} into \textbf{SELF-GRPO}, which introduces a secondary reward to improve consistency and stability across rollouts within the same sampling group. For each rollout $y_i \sim \pi_{\theta_\text{old}}(\cdot|x)$, the primary reward is $r_i = \mathcal{R}(y_i)$. We aggregate emotion distributions over all $n$ rollouts to form a core distribution $p^*$ over the top-3 emotions $\mathcal{E}_\text{top3}$:
\begin{equation}
\tilde{p}(e) = \frac{\sum_i p_i(e)}{\sum_{e'}\sum_i p_i(e')}, \quad
p^*(e) = \frac{\tilde{p}(e)}{\sum_{e'\in\mathcal{E}_\text{top3}}\tilde{p}(e')}
\end{equation}
The secondary reward counts how many of the top-3 consensus emotions appear in rollout $y_i$:
\begin{equation}
r_i^{(2)} = \frac{1}{3}\sum_{e\,\in\,\mathcal{E}_\text{top3}} \mathbf{1}[e \in Prediction_i]
\end{equation}

Both rewards are normalized. The combination weight $\lambda$ increases linearly with training progress to encourage early exploration and late consistency:
\begin{equation}
\lambda = \frac{t}{T}, \quad
A_i = \frac{r_i - \mu_r}{\sigma_r} + \lambda\,\frac{r_i^{(2)} - \mu_{r^{(2)}}}{\sigma_{r^{(2)}}}
\end{equation}
where $t$ is the current step and $T$ is the total training steps. The policy is updated by:
\begin{equation}
\label{loss_selfgrpo}
\mathcal{L}_{self\_grpo}(\theta) = \frac{1}{n}\sum_{i=1}^{n} \min\!\left(\rho_i A_i,\ \operatorname{clip}(\rho_i,\,1{-}\epsilon,\,1{+}\epsilon)\,A_i\right), \quad \rho_i = \frac{\pi_\theta(y_i\mid x)}{\pi_{\theta_\text{old}}(y_i\mid x)}
\end{equation}
\label{SELF-GRPO}

\subsection{SELF-EMO Training Algorithm}
Algorithm~\ref{alg:selfemo} summarizes the full SELF-EMO training process.
The framework starts from a cold-start SFT stage, then alternates between
two steps: SELF-GRPO updates the policy on the current buffer $\mathcal{B}$,
and the best rollout per prompt is filtered back into $\mathcal{B}$ to retrain
the base model. Each iteration feeds better data into the next, forming a
self-improving closed loop.

\begin{algorithm}[h]
\caption{SELF-EMO Closed-Loop Optimization}
\label{alg:selfemo}
\begin{algorithmic}[1]
\Require Original dataset $\mathcal{D}_0$; replay buffer $\mathcal{B} \leftarrow \mathcal{D}_0$;
         reward $\mathcal{R}(\cdot)$; iterations $K$; rollouts per prompt $n$
\Ensure Optimized policy $\pi_{\theta^*}$

\State \textbf{Cold Start:} train $\pi_\theta$ on $\mathcal{D}_0$ via supervised fine-tuning

\For{$k = 1$ to $K$}
    \State \textbf{// RL stage: rollout and policy update}
    \For{each prompt $x$ sampled from $\mathcal{B}$}
        \State Sample $n$ rollouts $\{y_i\}_{i=1}^n \sim \pi_\theta(\cdot \mid x)$
        \State Compute per-rollout rewards $r_i = \mathcal{R}(y_i)$ \hfill(Eq.~\ref{eq:reward})
        \State Update policy: $\theta \leftarrow \theta - \eta\,\nabla_\theta\,\mathcal{L}_{\text{SELF-GRPO}}$ \hfill(Eq.~\ref{loss_selfgrpo})
    \EndFor

    \State \textbf{// Data selection: best rollout per prompt into buffer}
    \For{each prompt $x$}
        \State $y^* = \arg\max_{i}\; r_i$
        \State Construct $d^* = (C_{new}, Label_{new})$ from $y^*$ \hfill(Eq.~\ref{eq:update})
        \State $\mathcal{B} \leftarrow \mathcal{B} \cup \{d^*\}$
    \EndFor

    \State \textbf{// Closed-loop retraining on updated buffer}
    \State Retrain base model $\pi_\theta$ on $\mathcal{B}$ via supervised fine-tuning
\EndFor

\State \Return $\pi_\theta$
\end{algorithmic}
\end{algorithm}
This design removes the need for manual data collection. By leveraging
ERC-related auxiliary tasks, the model generates and filters its own training
data throughout the process, keeping annotation costs low while maintaining
data quality via the reward-based selection in Eq.~\ref{eq:reward}.

\section{Experiments}
This section introduces the experimental setup, the experimental results, and an analysis of the results.
\subsection{Experimental Setting}
\noindent \textbf{Datasets.} Our methodology is grounded in three widely recognized benchmarks in the ERC domain: IEMOCAP ~\cite{iemocap}, MELD ~\cite{meld}, and EmoryNLP ~\cite{emorynlp}. These datasets serve as the seed data for persona initialization. Specifically, IEMOCAP consists of dyadic conversations among ten speakers (5,163/647/1,623 samples for train/dev/test). MELD and EmoryNLP, both derived from television series with multi-party dialogue settings, comprise 9,989/1,109/2,610 and 7,551/954/984 utterances, respectively. All model evaluations are strictly performed on the official test splits of these benchmarks.

\noindent \textbf{Baselines.} To verify the effectiveness of our approach, we compare it against prominent ERC methods, categorized into two groups: 
(1) \texttt{Traditional methods}, including DialogueRNN ~\cite{diarnn}, ICON ~\cite{hazarika-etal-2018-icon}, DialogueGCN ~\cite{ghosal2019dialoguegcngraphconvolutionalneural}, COSMIC ~\cite{ghosal2020cosmiccommonsenseknowledgeemotion}, MMGCN ~\cite{hu2021mmgcnmultimodalfusiondeep}, DAG-ERC ~\cite{shen2021directedacyclicgraphnetwork}, LR-GCN ~\cite{9556142}, Multi-DAG+CL ~\cite{nguyen2024curriculumlearningmeetsdirected}, CBERL ~\cite{meng2023deepimbalancedlearningmultimodal}, DER-GCN ~\cite{ai2024dergcndialogueeventrelationaware}, and LSDGNN+ICL \cite{Li_2025}; 
(2) \texttt{LLM-based methods}, such as InstructERC ~\cite{lei2024instructercreformingemotionrecognition}, BiosERC ~\cite{Xue_2024}, and PRC-EMO ~\cite{li2025llmsfeelteachingemotion}.

\noindent \textbf{Implementation Details.} The experiments are conducted on a cluster of eight NVIDIA H100 GPUs, utilizing Qwen3-4B and Qwen3-8B ~\cite{qwen3} as the primary backbone models. Departing from the common practice where models are restricted to single-dataset training—a setup prone to overfitting and limited generalization—this work adopts a more robust strategy. Specifically, a single unified model is trained across all three ERC training sets and subsequently evaluated on each official test split. For a rigorous and fair comparison, the two most competitive LLM-based baselines are retrained under this identical multi-dataset setting. Additional hyperparameter configurations and training details are provided in Appendix \ref{appendex:parameters}.

\subsection{Main results}
As shown in Table \ref{tab:main}, under the unified training setting, existing LLM-based methods exhibit limited generalization across different conversation contexts. In contrast, the proposed SELF-EMO framework achieves substantial performance gains. Specifically, our model realizes an accuracy  increase of \textbf{+6.33\%} on the Qwen3-4B backbone and \textbf{+8.54\%} on the Qwen3-8B backbone, with consistent improvements observed across all benchmarks. These results demonstrate the robust generalization of SELF-EMO and empirically validate the effectiveness of the self-evolution paradigm in enhancing emotional intelligence through autonomous agent interactions.

\begin{table}[t]

\centering
\renewcommand{\arraystretch}{1.25}
\resizebox{\linewidth}{!}{
\begin{tabular}{llclcccccccc}
\hline
                                                         &                                     &                                                                                     & \multicolumn{2}{c}{\textbf{IEMOCAP}} & \multicolumn{2}{c}{\textbf{MELD}} & \multicolumn{2}{c}{\textbf{EMORYNLP}} & \multicolumn{2}{c}{\textbf{AVG}} \\ \cline{4-11} 
\multirow{-2}{*}{\textbf{Method}}                        & \multirow{-2}{*}{\textbf{Backbone}} & \multirow{-2}{*}{\textbf{\begin{tabular}[c]{@{}c@{}}External \\ FREE\end{tabular}}} & Acc.              & W-F1             & Acc.            & W-F1            & Acc.              & W-F1              & Acc.            & W-F1           \\ \hline \hline
\multicolumn{11}{c}{Dataset-Specific Setting}                                                                                                                                                                                                                                                                                              \\ \hline \hline
\multicolumn{11}{l}{\texttt{LLM-based methods}}                                                                                                                                                                                                                                                                                                     \\
DialogueRNN                                              & GRU                                   & -                                                                                   & 63.40             & 62.75            & 60.27           & 57.03           & -                 & 31.70                 & -               & 50.49              \\
ICON                                                     & -                                   & -                                                                                   & 64.00             & 63.50            & -               & 56.30           & -                 & -                 & -               & -              \\
DialogueGCN                                              & -                                   & -                                                                                   & 65.25             & 64.18            & -               & 58.10           & -                 & -                 & -               & -              \\
COSMIC                                                   & -                                   & -                                                                                   & -                 & 65.28            & -               & 65.21           & -                 & 38.11                 & -               & 56.20              \\
MMGCN                                                    & -                                   & -                                                                                   & -                 & 66.22            & -               & 58.65           & -                 & -                 & -               & -              \\
DAG-ERC                                                  & -                                   & -                                                                                   & 67.53             & 68.03            & 63.98           & 63.63           & -                 & 39.02                 & -               & 56.89              \\
LR-GCN                                                   & -                                   & -                                                                                   & 68.50             & 68.30            & -               & 65.60           & -                 & -                 & -               & -              \\
MultiDAG+CL                                              & -                                   & -                                                                                   & 69.11             & 69.08            & -               & 64.00           & -                 & -                 & -               & -              \\
CBERL                                                    & -                                   & -                                                                                   & 69.36             & 69.27            & 67.78           & 66.89           & -                 & -                 & -               & -              \\
DER-GCN                                                  & -                                   & -                                                                                   & 69.70             & 69.40            & 66.80           & 66.10           & -                 & -                 & -               & -              \\
LSDGNN+ICL                                               & -                                   & -                                                                                   & 70.35             & 70.24            & 64.67           & 64.07           & -                 & -                 & -               & -              \\
\multicolumn{11}{l}{\texttt{Traditional methods}}                                                                                                                                                                                                                                                                                                   \\
InstructERC                                              & LLaMA2-7B                           & NO                                                                                  & -                 & 71.39            & -               & 69.15           & -                 & 41.39                 & -               & -              \\
BiosERC                                                  & LLaMA2-13B                          & NO                                                                                  & -                 & 71.19            & -               & 69.83           & -                 & 41.68                 & -               & -              \\
PRC-Emo                                                  & Qwen2.5/3                       & NO                                                                                  & 71.03             & 71.95            & 71.50           & 70.44           & -                 & -                 & -               & -              \\ \hline \hline
\multicolumn{11}{c}{Dataset-Unified Setting}                                                                                                                                                                                                                                                                                               \\ \hline \hline
Seed-1.8                                                  & Seed-1.8                          &YES                                                                                  & 18.30             & 18.54            & 26.82           & 25.31           & 21.85             & 20.99             & 22.32               & 21.61              \\
DeepSeek-V3.2                                                  & DeepSeek-V3.2                          &YES                                                                                  & 19.90             & 19.84            & 20.69           & 20.57           & 19.82             & 18.55             & 20.13               & 19.65              \\
Qwen3-4B                                                  & Qwen3-4B                          &YES                                                                                  & 56.07             & 55.83            & 60.96           & 59.55           & 41.87             & 41.50             & 52.96               & 52.29              \\
Qwen3-8B                                                  & Qwen3-8B                          &YES                                                                                  & 57.67             & 57.33            & 59.43           & 59.21           & 43.19             & 42.71             & 53.43               & 53.08              \\
BiosERC                                                  & LLaMA2-13B                          & NO                                                                                  & 28.53             & 28.84            & 62.13           & 61.59           & 19.83             & 20.55             & 36.83               & 36.99              \\
PRC-Emo                                                  & Qwen3-8B                            & NO                                                                                  & 27.39             & 27.53            & 64.07           & 63.88           & 22.53             & 21.94             & 37.99               & 37.78              \\ \hline
\rowcolor[HTML]{EFEFEF} 
\cellcolor[HTML]{EFEFEF}                                 & Qwen3-4B                            & YES                                                                                 & \textbf{64.60}    & \textbf{63.57}   & \textbf{67.74}  & \textbf{67.32}  & \textbf{45.52}    & \textbf{44.98}    & \textbf{59.29}               & \textbf{58.62}              \\
\rowcolor[HTML]{EFEFEF} 
\multirow{-2}{*}{\cellcolor[HTML]{EFEFEF}SELF-EMO(Ours)} & Qwen3-8B                            & YES                                                                                 & \textbf{66.11}    & \textbf{64.81}   & \textbf{71.92}  & \textbf{70.30}  & \textbf{47.87}    & \textbf{47.61}    & \textbf{61.97}               & \textbf{60.91}              \\ \hline
\end{tabular}
}
\caption{Performance comparison on IEMOCAP, MELD and EmoryNLP. \textbf{Dataset-Specific Setting}: trained and tested on each dataset separately. \textbf{Dataset-Unified Setting}: trained on combined three datasets and tested on each official split. Our SELF-EMO achieves consistent gains across all benchmarks.}
\label{tab:main}
\end{table}

\subsection{Ablation Experiments}
We conduct ablation experiments on SELF-EMO-4B to assess each component's contribution (Table \ref{tab:ablation}). Regarding the data produced by SELF-EMO, w/ COLD uses the cold-start data generated by the expert model to first perform SFT, and then uses the resulting model as the base model for the RL training stage. w/o SELF-GRPO indicates that we directly apply the vanilla GRPO algorithm for RL-stage fine-tuning. w/o SELF-EMO indicates that we train with GRPO using all data from the original datasets. 
\definecolor{darkred}{RGB}{160,0,0}
\definecolor{darkgreen}{RGB}{0,120,0}
\begin{table}[h]
\renewcommand{\arraystretch}{1.2}
\resizebox{\linewidth}{!}{
\begin{tabular}{lllllll}
\hline
\multirow{2}{*}{Model} & \multicolumn{2}{c}{\begin{tabular}[c]{@{}c@{}}Data \\ Composition\end{tabular}} & \multicolumn{1}{c}{\multirow{2}{*}{EMORYNLP}} & \multicolumn{1}{c}{\multirow{2}{*}{IEMOCAP}} & \multicolumn{1}{c}{\multirow{2}{*}{MELD}} & \multicolumn{1}{c}{\multirow{2}{*}{AVG}} \\ \cline{2-3}
                       & \multicolumn{1}{c}{synthetic}           & \multicolumn{1}{c}{original}          & \multicolumn{1}{c}{}                          & \multicolumn{1}{c}{}                         & \multicolumn{1}{c}{}                      & \multicolumn{1}{c}{}                     \\ \hline
SELF-EMO     & 8k & 15k    & 45.52 & 64.60 & 67.74 & 59.29 \\

w/ COLD     & 13k  & 15k    & 42.46{\color{darkgreen}(-3.06)} 
                        & 61.95{\color{darkgreen}(-2.65)} 
                        & 67.03{\color{darkgreen}(-0.71)} 
                        & 57.15{\color{darkgreen}(-2.14)} \\
                        
w/o SELF-GRPO & 8k  & 15k & 42.23{\color{darkgreen}(-3.29)} 
                        & 61.10{\color{darkgreen}(-3.5)} 
                        & 65.14{\color{darkgreen}(-2.6)} 
                        & 56.15{\color{darkgreen}(-3.14)} \\
                        
w/o SELF-EMO       & 0k  & 22.5k    & 42.07{\color{darkgreen}(-3.45)} 
                        & 58.15{\color{darkgreen}(-6.45)} 
                        & 62.13{\color{darkgreen}(-5.61)} 
                        & 54.11{\color{darkgreen}(-5.18)} \\
 \hline
\end{tabular}}
\caption{Ablation experiments on each component. }

\label{tab:ablation}
\end{table}
The experiments demonstrate that our method does not rely on data produced by the expert model. Instead, during the cold-start phase, the model only learns the corresponding reasoning and output patterns, and incorporating expert-generated data for SFT actually degrades model performance. SELF-GRPO enables the model to perform macro-level reward modeling over sampled outputs, leading to more stable emotional understanding outputs and consequently improved performance. Overall, both the self-play data generated by SELF-EMO and the training algorithm are crucial to the model's performance gains; applying GRPO fine-tuning with only the original data leads to a sharp average performance drop of 5.18\%.

\subsection{Analysis}
\label{analysis}
\begin{figure}[h]
\centering
\begin{subfigure}{0.49\linewidth}
    \centering
    \includegraphics[width=\linewidth]{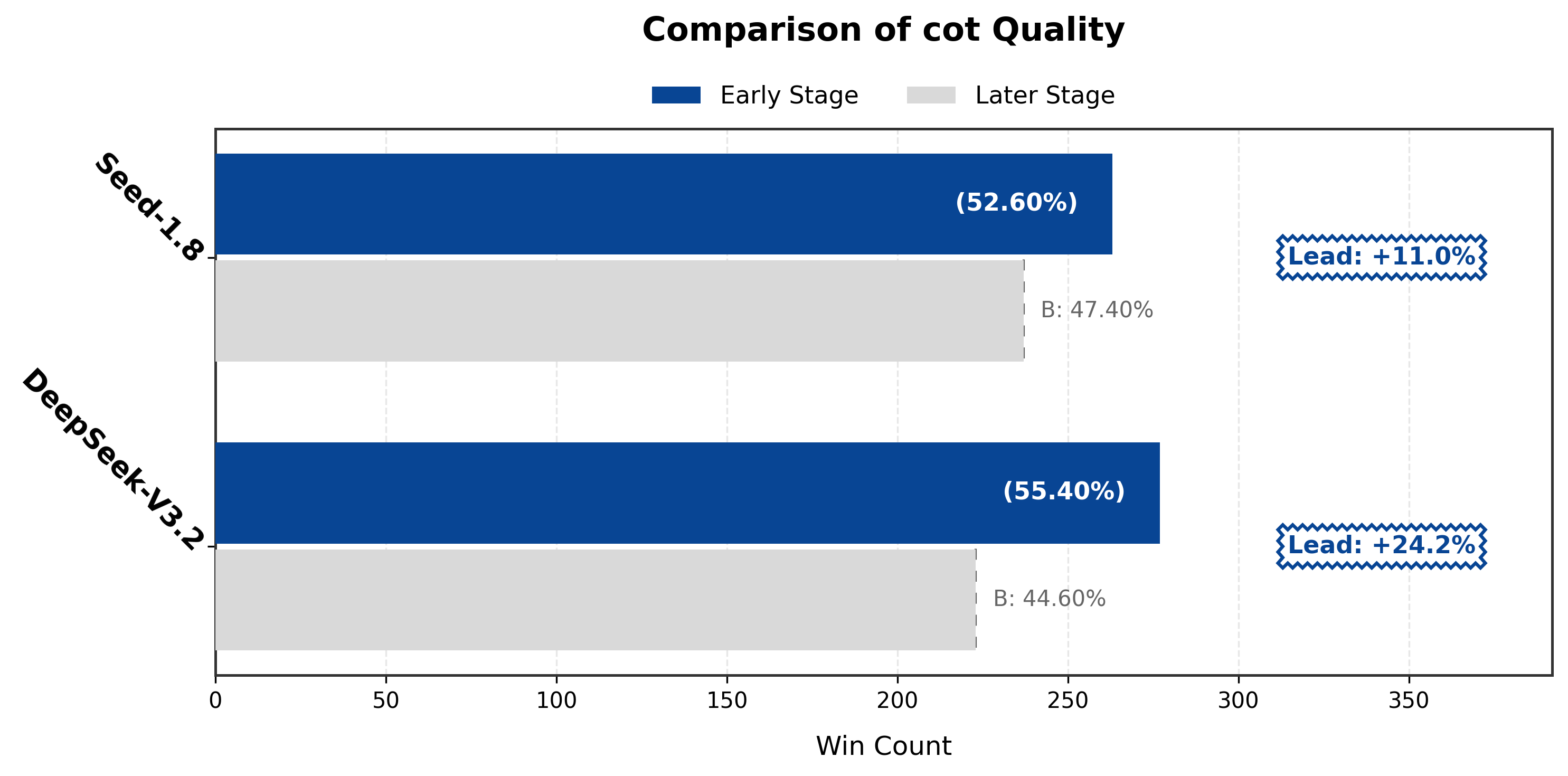}
    \caption{Self-personality understanding}
    \label{fig:sub1}
\end{subfigure}
\hfill
\begin{subfigure}{0.49\linewidth}
    \centering
    \includegraphics[width=\linewidth]{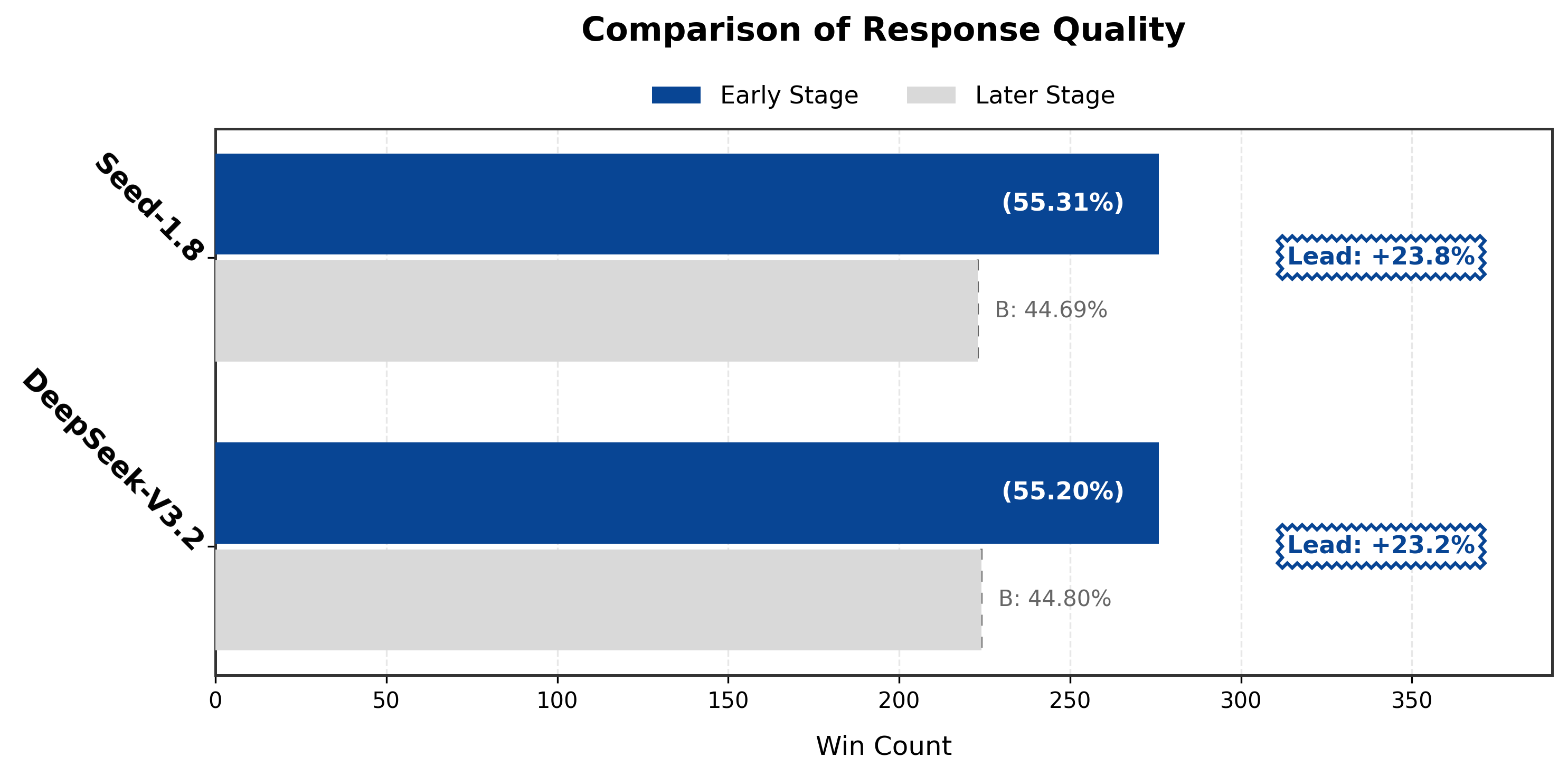}
    \caption{Self-emotion expression}
    \label{fig:sub2}
\end{subfigure}

\caption{Performance improvement on two self-supervised tasks. DeepSeek and Seed serve as judges for A/B testing on generated responses. SELF-EMO significantly boosts response quality after training.}
\label{fig:two_subtasks}
\end{figure}
\subsubsection{Performance Improvement on Self-supervised Tasks}

As shown in Fig.~\ref{fig:two_subtasks}, we collect online sampled examples from models at the early stage of RL training and the later stage of training, and randomly sample them for analysis. We use two expert models, Seed-1.8 and DeepSeek-V3.2, to evaluate the performance improvement of our method on the two unsupervised tasks. For the same problem, the outputs of the two models are simultaneously given to the expert models for an AB-test, where the experts are required to determine which output is better.

The results show that both in self-personality understanding (a) and self-emotion expression (b), the model at the later stage of training is clearly better than the model at the early stage. This directly demonstrates the effectiveness of our unsupervised method: by explicitly optimizing the ERC task, the other two tasks can also be optimized in an unsupervised manner. More analysis and observations on task entropy reduction are provided in Appendix \ref{appendix:train}, and specific examples can be found in Appendix \ref{appendix:case}.

\subsubsection{Impact of Personality Traits on Reward Acquisition}
To further analyze the influence of personality traits on obtaining IOU-based prediction rewards, we select the five most frequent personality categories from the clustering results for analysis. As shown in Table~\ref{tab:personality_distribution}, the column abbreviations correspond to the following traits: CPDE denotes \texttt{composed\_practical\_direct\_efficient}, EEIPRC denotes \texttt{emotional\_expressiveness\_interpersonal\_playfulness\_reactive\_curiosity}, ERDCPD denotes \texttt{emotional\_reactivity\_defensive\_confrontation\_pragmatic\_dominance}, EDAE denotes \texttt{emotional\_directness\_assertive\_engagement}, and HRS denotes \texttt{humor\_regulated\_social}.
\begin{table}[h]
\renewcommand{\arraystretch}{1.2}
\resizebox{\linewidth}{!}{
\centering
\small
\begin{tabular}{lcccccc}
\hline
Model & CPDE & EEIPRC & ERDCPD & EDAE & HRS & AVG \\
\hline
Seed-1.8 & 13.06\% & 12.81\% & 13.03\% & 14.21\% & 15.75\% & 13.77\% \\
DeepSeek-V3.2 & 10.25\% & 10.89\% & 9.64\% & 9.21\% & 10.21\% & 10.04\% \\
SELF-EMO-8B & 57.09\% & 58.83\% & 58.16\% & 56.32\% & 61.98\% & 58.48\% \\
\hline
\end{tabular}}
\caption{Distribution of personality traits predicted by different models. SELF-EMO obtains significantly higher rewards across all personality types.}
\label{tab:personality_distribution}
\end{table}
By analyzing Table~\ref{tab:personality_distribution}, we observe that the optimized SELF-EMO achieves significantly higher rewards than the two larger models. The calm and objective trait HRS consistently yields higher rewards, while the more emotional and confrontational ERDCPD results in lower rewards, indicating that personality traits influence emotion recognition. More details on training data composition are provided in Appendix \ref{appendex:data}.

\section{Conclusion}
In conclusion, we present SELF-EMO, a psychologically motivated reinforcement learning framework that leverages three strongly correlated tasks for collaborative entropy reduction. SELF-EMO achieves state-of-the-art ERC performance under the Dataset-Unified Setting while significantly reducing inference overhead, and the data produced during training helps alleviate the scarcity of high-quality annotations in this domain.

\bibliographystyle{unsrt}
\bibliography{main.bib}
\newpage
\appendix
\section{Prompts}
\label{appendix:prompts}
In this section, we provide the prompts given to the expert models during the data cold-start stage, including the scenario\&personality extraction prompt(Table\ref{tab:emotion_prompt}) and the prompt used to guide the generation of SFT data(Table\ref{tab:sft_prompt}). These prompts play an important role in enabling the model to learn the output format used in the self-play stage.
\begin{table}[h]
\centering
\begin{tcolorbox}[title=\textbf{Scenario and Personality Information Extraction}]

You are a professional emotional analyst and personality profiling expert.

Your task is to analyze a given multi-speaker dialogue and produce a distinct personality profile for each participant, as well as extract the overall theme of the conversation.

Please strictly follow the requirements below.

\textbf{1. Analysis Dimensions} (Inference-based, no fabrication)\\
For each speaker, infer their characteristics ONLY based on the dialogue content. The analysis should consider (but is not limited to) the following aspects:

1. Personality traits (e.g., rational vs. emotional, introverted vs. extroverted, cautious vs. impulsive, gentle vs. assertive)
2. Estimated age range (use vague ranges such as "20–30s", "30+"; do NOT provide a specific age)
3. Possible gender (if unclear, explicitly state "gender not obvious")
4. Likely education or cognitive background (e.g., higher-education background, practical/non-academic thinking style)
5. Speaking and expression style (e.g., formal vs. casual, logic-driven vs. emotion-driven, restrained vs. expressive wording)
6. Emotional state and stance (e.g., anxious, complaining, calmly analyzing, comforting others, confrontational vs. cooperative)

Requirements:
- All conclusions must be grounded in the dialogue itself; do NOT invent information.
- Each character's personality profile must be clearly distinctive.
- Avoid generic, vague, or template-like descriptions (e.g., "quite normal", "relatively rational").

\textbf{2. Personality Summary Requirement}
- For each speaker, compress the analysis into ONE concise but information-dense sentence.
- The sentence should naturally integrate multiple dimensions rather than listing attributes mechanically.

Example style (for reference only, do NOT copy):
"The speaker appears to be a cautious and rational individual, likely in their 30s, with a higher-education background, expressing themselves in a structured and emotionally restrained manner."

\textbf{3. Theme Extraction}
- Abstract one or more core themes from the entire dialogue.
- Themes should be semantic-level summaries, not surface-level restatements.
- If conflicts or disagreements exist, reflect them in the theme when appropriate.

\textbf{4. Output Format (Critical)}
After completing the analysis, output  a Python code block containing a dictionary in the following format:

... analysis...
\begin{verbatim}
{
    "Speaker A Name": "One-sentence personality profile",
    "Speaker B Name": "One-sentence personality profile",
    ...
    "Theme": "Core theme(s) of the conversation"
}
\end{verbatim}
Output rules:
You should first analyze each speaker, and then output the Pyhon code block.

Do NOT output anything outside the Python code block.

Speaker names must exactly match those used in the dialogue.

\end{tcolorbox}

\caption{Prompt used for scenario and personality information extraction.}
\label{tab:emotion_prompt}
\end{table}

\begin{table}[h]
\centering
\begin{tcolorbox}[title=\textbf{Prompt for SFT Data Generation}]
\textbf{System Prompt}\\ 
You are now a person named ``\{name\}'' with the following characteristics: \{role\_message\}. 
You are currently engaged in a conversation with others.

You should:

(1) Infer the emotion, personality traits, and underlying thoughts of the last speaker in the conversation history, and provide an emotional profile (as a dictionary) for that person.

(2) Reflect based on your own role persona, then formulate your response—along with the emotions embedded in your reply.

\{emo\_list\}
\\

\textbf{User Prompt}\\ 
\{history\}

You now need to analyze the conversation, provide an analysis of the last utterance along with your own reasoning, and package them into a Python dictionary.
\\

Assistant Output Format 
\begin{verbatim}
<think>
... reasoning process ...
</think>

{
 "last_emotions": ...,
 "my_emotions": ...,
 "my_output": ...
}
\end{verbatim}
\end{tcolorbox}

\caption{Prompt for SFT Data Generation}
\label{tab:sft_prompt}
\end{table}

\section{Data Analysis}
\label{appendex:data}
We apply HDBSCAN to automatically cluster the extracted personality descriptions. To avoid potential bias in the clustering process, the original personality names are masked before clustering. In total, 74 clusters are obtained.

\begin{figure}[t]
    \centering
    \includegraphics[width=1.0\linewidth]{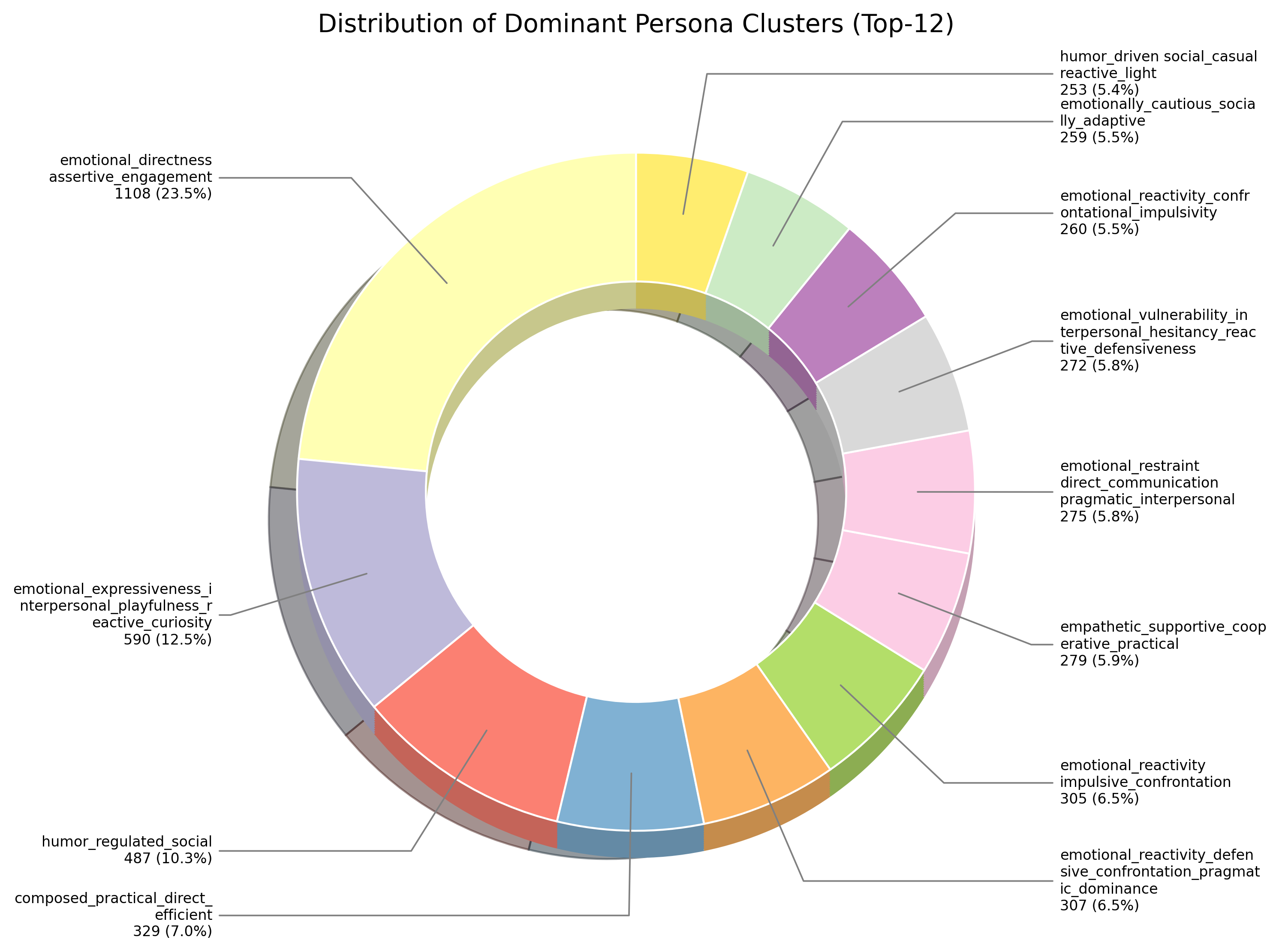}
    \caption{Personality Category.}
    \label{fig:personal}
\end{figure}

As shown in Fig.~\ref{fig:personal}, we visualize the 12 clusters with the largest number of samples. Tables~\ref{tab:personality_distribution_1} and \ref{tab:personality_distribution_2} present the specific personality category names and their corresponding counts.

It is worth noting that, due to the nature of the clustering algorithm, some clusters may contain semantically similar descriptions.

\begin{table}[h]
\centering
\small
\renewcommand{\arraystretch}{1.2}
\begin{tabularx}{\linewidth}{X c X c}
\hline
Type & Num & Type & Num \\
\hline \hline
emotional directness assertive engagement & 1108 & emotional expressiveness interpersonal playfulness reactive curiosity & 590 \\ 
humor regulated social & 487 & composed practical direct efficient & 329 \\
emotional reactivity defensive confrontation pragmatic dominance & 307 & emotional reactivity impulsive confrontation & 305 \\
empathetic supportive cooperative practical & 279 & emotional restraint direct communication pragmatic interpersonal & 275 \\
emotional vulnerability interpersonal hesitancy reactive defensiveness & 272 & emotional reactivity confrontational impulsivity & 260 \\
emotionally cautious socially adaptive & 259 & humor driven social casual reactive light & 253 \\
direct emotional pragmatic & 210 & calm cooperative pragmatic & 210 \\
emotionally regulated socially engaged assertive receptive & 201 & humor driven social reactivity & 194 \\
expressive playful adaptable supportive & 193 & emotional neutrality direct factuality & 189 \\
reactive defensive unfiltered & 181 & assertive emotionally direct autonomous & 174 \\
emotionally reactive assertive confrontational & 171 & emotional reactivity interpersonal hesitancy anxiety driven & 166 \\
cautious empathetic guarded & 149 & emotional mixed interpersonal casual pragmatic hesitant & 137 \\
procedural calm solution oriented & 133 & emotional expressivity interpersonal reactivity impulsive coping & 129 \\
anxious evasive overthinker & 129 & emotional vulnerability reactive interpersonal seek withdraw defensive sensitivity & 127 \\
pragmatic calm direct supportive & 123 & emotionally restrained pragmatic supportive & 109 \\
reserved practical & 104 & emotional restraint passive compliance & 103 \\
reactive urgent assertive pragmatic & 102 & direct pragmatic emotional control & 102 \\
trauma bound guilt moral compulsion & 100 & emotional reactivity assertive adaptability & 98 \\
pragmatic defensiveness & 93 & emotional reactivity interpersonal evasiveness & 92 \\
humor driven socially adaptive playful critique & 90 & rational calm procedural neutral & 86 \\
\hline \hline
\end{tabularx}
\caption{Personality type distribution (Part 1).}
\label{tab:personality_distribution_1}
\end{table}

\begin{table}[h]
\centering
\small
\renewcommand{\arraystretch}{1.2}
\begin{tabularx}{\linewidth}{X c X c}
\hline
Type & Num & Type & Num \\
\hline \hline
emotional reactivity spontaneous expression direct engagement curiosity driven & 84 & emotional reactivity confrontational impulsive & 83 \\
calm restraint polite cooperative rule oriented & 81 & emotional expressiveness persistent engagement nature reactivity idealistic connection & 80 \\
emotionally expressive socially playful & 78 & emotional reactivity defensive defiance persistent worthlessness validation struggle & 78 \\
emotional vulnerability reactive grief empathetic support resilient reframing & 76 & nostalgic guarded withdrawal & 76 \\
calm direct practical & 76 & empathetic supportive pragmatic calm & 74 \\
high energy expressive cooperative optimistic & 72 & desperate urgency hope clinging plea driven defensive irrationality & 70 \\
emotional reactivity defensive pessimism & 68 & emotional fragility grief reactivity & 67 \\
empathetic pragmatic supportive patient & 66 & calm procedural assertive & 65 \\
emotionally restrained casual deescalator & 62 & empathetic supportive practical & 62 \\
emotionally expressive playfully spontaneous socially engaged & 60 & empathetic listener & 59 \\
pragmatic detached skeptical & 59 & pragmatic emotionally restrained direct task oriented & 59 \\
emotional volatility reactive sensitivity & 59 & warmth driven expressive reactive & 58 \\
emotional reactivity assertive defensiveness interpersonal fluctuation & 58 & emotional reactivity expressive friendliness pragmatic cooperativeness & 58 \\
emotional reactivity impulsive seeking connection conflict prone & 57 & calm cooperative professional & 55 \\
pragmatic direct & 55 & socially oriented curious engaging & 52 \\
humor driven social engager & 51 & rational casual defensive fact oriented & 51 \\
social playfulness & 50 & curious neutral direct inquisitive & 50 \\
\hline \hline
\end{tabularx}
\caption{Personality type distribution (Part 2).}
\label{tab:personality_distribution_2}
\end{table}
\newpage

\section{Multi-task Training Metrics and Analysis}
\label{appendix:train}
\subsection{Consistent Growth of Reward}
\begin{figure}[h]
\centering
\begin{subfigure}{0.49\linewidth}
    \centering
    \includegraphics[width=\linewidth]{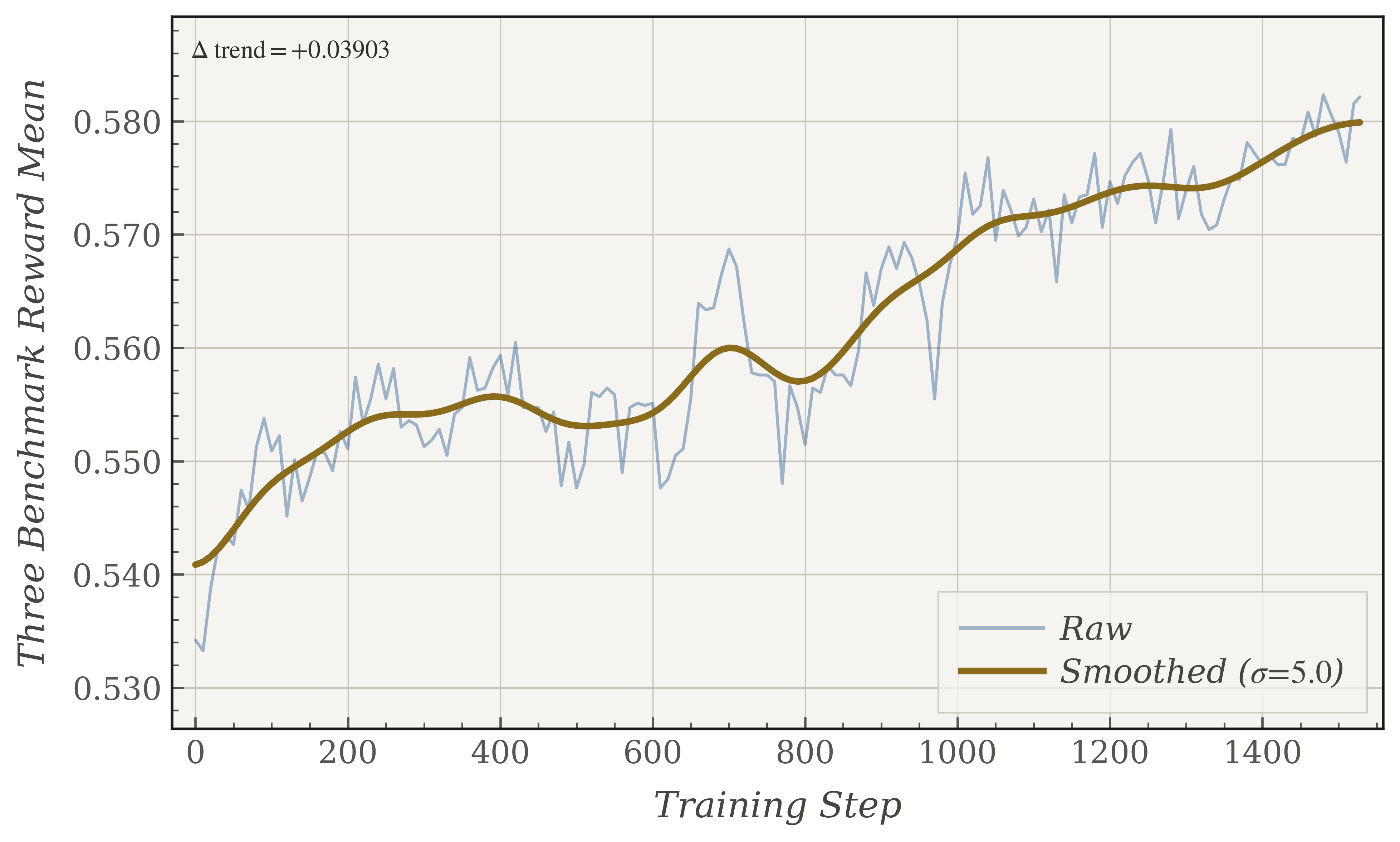}
    \caption{Acc of three emo benchmark.}
    \label{fig:reward_sub1}
\end{subfigure}
\hfill
\begin{subfigure}{0.49\linewidth}
    \centering
    \includegraphics[width=\linewidth]{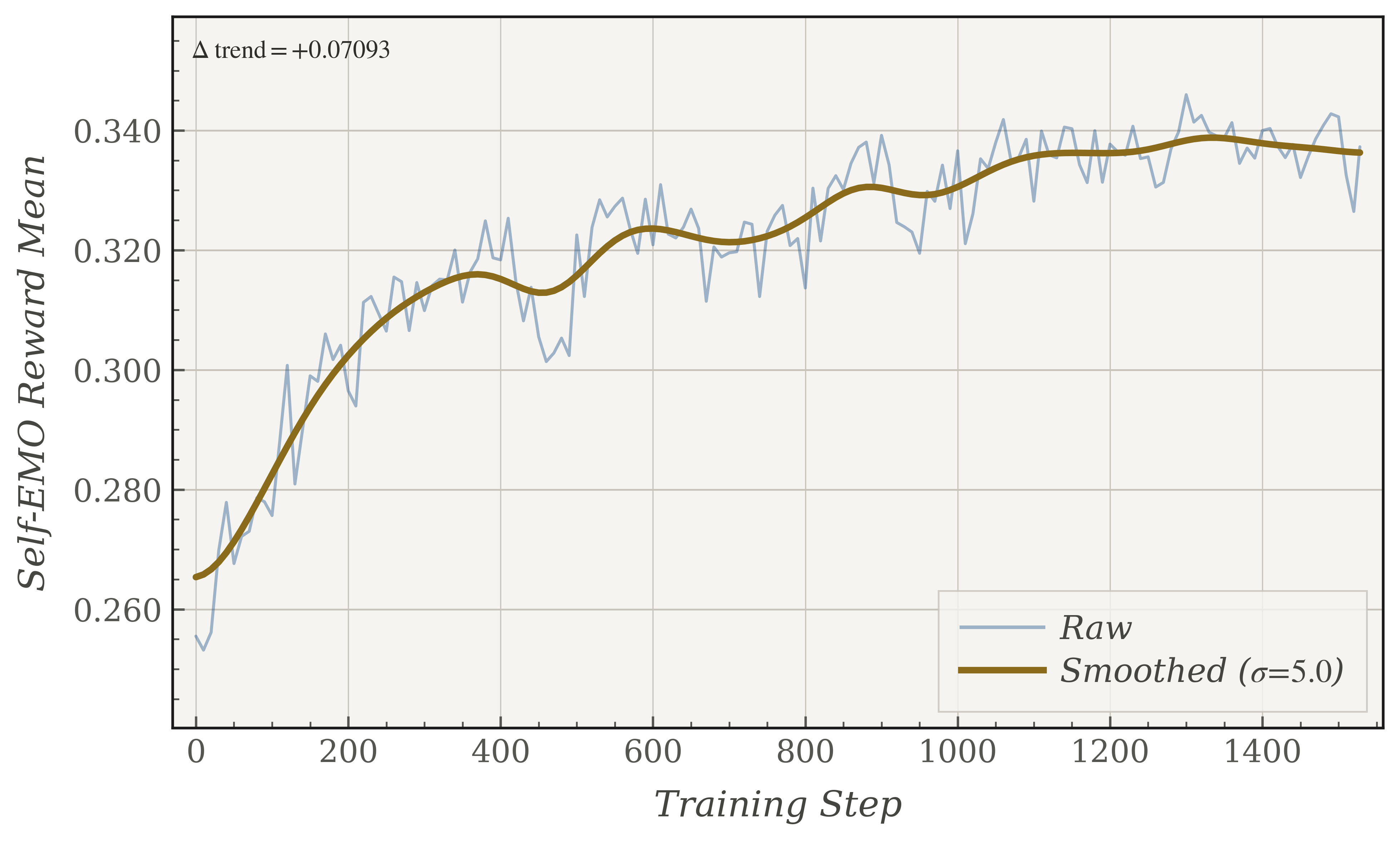}
    \caption{Self-emotion IOU reward}
    \label{fig:reward_sub2}
\end{subfigure}

\caption{Steady Increase in Reward.}
\label{fig:reward_increase}
\end{figure}

As shown in Fig.~\ref{fig:reward_increase}, under our experimental settings, SELF-EMO maintains a steady increase in reward throughout training without experiencing collapse, across both three public benchmarks (IEMOCAP, MELD, EmoryNLP) and the SELF-EMO split test set.

Under different reward designs, compared to the reward curves on public datasets based on accuracy (Fig.\ref{fig:reward_sub1}), the IOU-based reward curve (Fig.\ref{fig:reward_sub2}) is smoother and more stable. This demonstrates that the proposed IOU reward is not only more robust and generalizable, but also plays a key role in stabilizing training.

\newpage
\subsection{Joint Entropy Reduction in Multi-task Learning}
\begin{figure}[h]
\centering
\begin{subfigure}{0.49\linewidth}
    \centering
    \includegraphics[width=\linewidth]{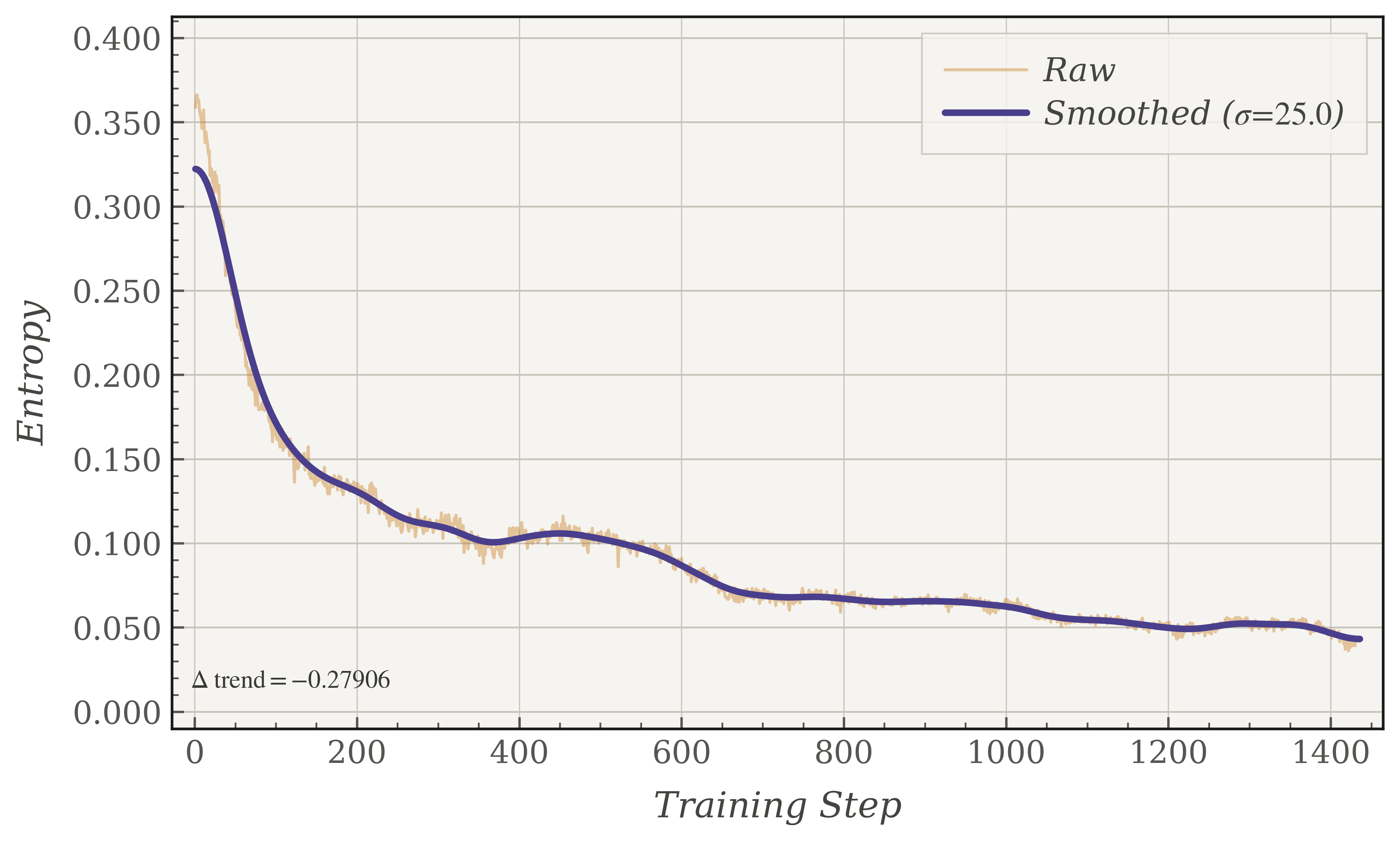}
    \caption{Entropy dynamics of the policy model}
    \label{fig:entropy_sub1}
\end{subfigure}
\hfill
\begin{subfigure}{0.49\linewidth}
    \centering
    \includegraphics[width=\linewidth]{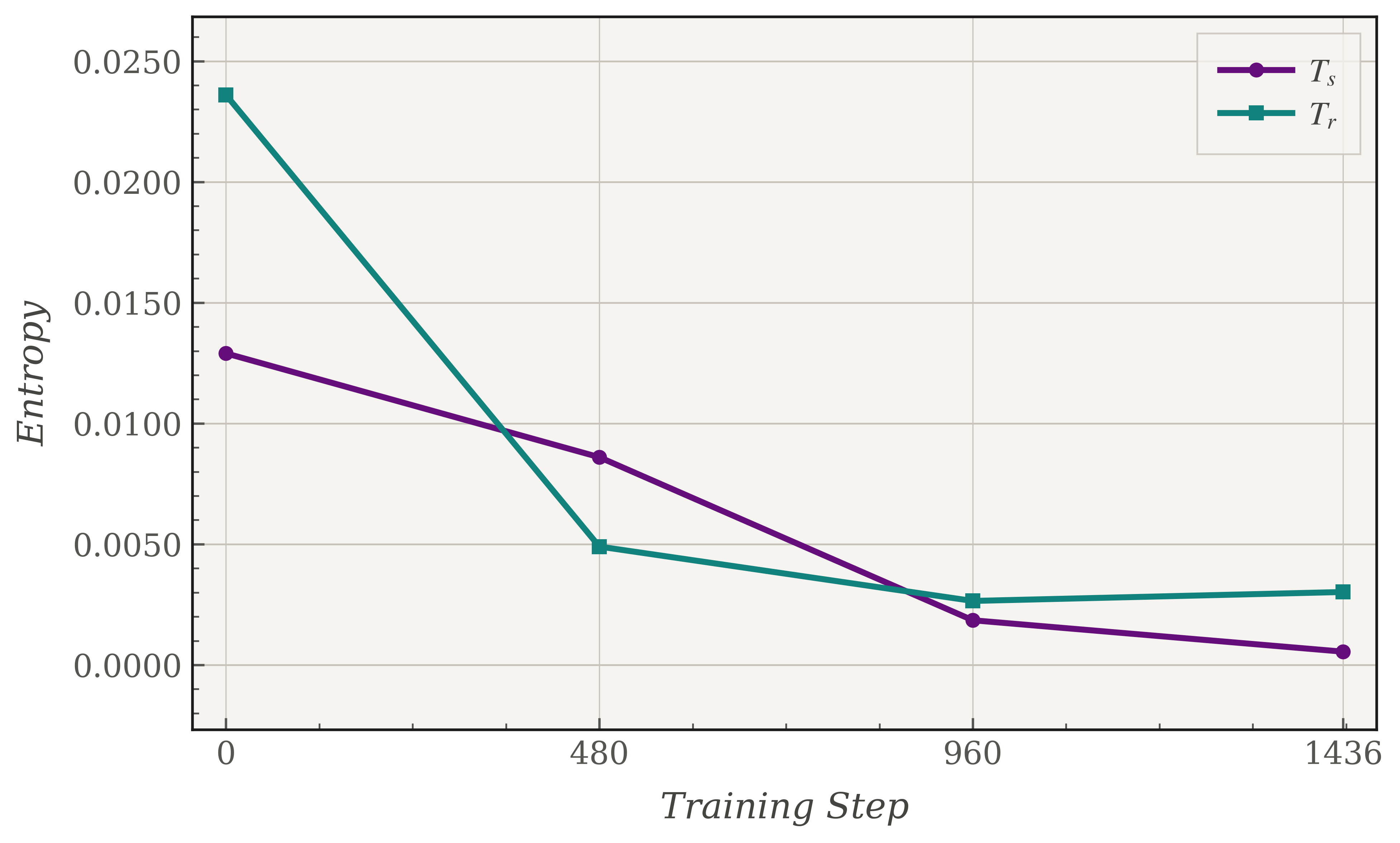}
    \caption{Entropy reduction in Self-Emotion tasks}
    \label{fig:entropy_sub2}
\end{subfigure}

\caption{Entropy reduction dynamics during multi-task training.}
\label{fig:entropy_down}
\end{figure}

To further substantiate the effectiveness of the proposed multi-task collaborative entropy reduction, we track the entropy of the policy model throughout training. As shown in Fig.~\ref{fig:entropy_sub1}, the entropy steadily decreases over time, with a notably sharper decline in the early stages. This trend is consistent with Fig.~\ref{fig:reward_increase}, where the reward improves most rapidly at the beginning of training, highlighting the alignment between reinforcement learning dynamics and entropy reduction.

In addition, we measure the entropy of the model on the two proposed tasks: self-emotion generation $T_s$ and self-emotion expression $T_e$ after each training epoch. Concretely, each checkpoint is evaluated on the SELF-EMO test split, and the average token-level entropy is computed for each task. As illustrated in Fig.~\ref{fig:entropy_sub2}, both tasks exhibit a clear downward trend in entropy during training, again with a more pronounced decrease in the early phase. This consistent pattern across tasks supports the collaborative entropy reduction hypothesis and suggests that supervised fine-tuning on the ERC task effectively promotes performance gains in both self-emotion generation and self-emotion expression.

\newpage
\section{Response Comparison Before and After Training}
\label{appendix:case}
As described in Section~\ref{analysis}, we conduct our analysis using samples generated by the model at the early stage of training and at the later stage of training. We then adopt an \texttt{LLM-as-a-judge} evaluation framework to compare the performance improvement of the model on the tasks of personality-conditioned self-emotion generation and self-emotion expression. A concrete example is shown in Table~\ref{tab:comparison_1}.

In this task, the model is required to first perform reasoning in the chain-of-thought process: 
(1) analyze the emotional state of the last speaker in the conversation, 
(2) generate its own emotional state based on the provided personality description and the current context, and 
(3) produce a response that is consistent with both the contextual situation and the specified personality, grounded in the analysis of others' emotions and its own generated emotional state.

When applying the \texttt{LLM-as-a-judge} protocol, we instruct the expert model Seed-1.8 to conduct the evaluation from five aspects: the personality description, the contextual situation, the analysis of \texttt{answer1}, the analysis of \texttt{answer2}, and the final comparison between \texttt{answer1} and \texttt{answer2}. During the evaluation process, we do not disclose any additional information about the identities of \texttt{answer1} and \texttt{answer2}, ensuring that the expert model’s judgment is not influenced by the source of the responses.

As shown in Table~\ref{tab:judge_1}, the expert model concludes that \texttt{answer1} is the superior response under the given personality and contextual constraints. In contrast, \texttt{answer2} neither reflects the personality trait of a ``casual, socially proactive style'' nor fits appropriately within the current conversational context.

To highlight the differences between the early and later stages of training, we also provide additional examples as shown in Table\ref{tab:comparison_2} and Table\ref{tab:judge_2}.
\begin{table}[h]
\centering
\small
\begin{tcolorbox}[title=\textbf{Response Comparison Example 1 (GOLD-neutral:0.85, surprise:0.45, fear:0.25)}, colback=white, colframe=black]

\begin{tabular}{p{0.96\linewidth}}

\textbf{SYSTEM} \\
You are now a person named Monica with the following characteristics: An expressive and socially proactive individual, likely in her 20s–30s, who uses casual, engaging questions to foster group connection and share emotional observations.        

You are currently engaged in a conversation with others.  

\#\#\# Available emotion labels: neutral, surprise, fear, sadness, joy, disgust, anger.\\

\textbf{USER} \\
\#\#\# Given the following conversation as a context

Monica: He is so cute.  So, where did you guys grow up?  

Angela: Brooklyn Heights.\quad Bob: Cleveland.

Monica: How, how did that happen?

Joey: Oh my god.\quad\quad\quad\quad\quad Monica: What?

Joey: I suddenly had the feeling that I was falling. But I'm not.

You now need to analyze the conversation, provide an analysis of the last utterance along with your own reasoning, and package them into a Python dictionary.

\end{tabular}

\begin{tabular}{p{0.47\linewidth}|p{0.47\linewidth}}

\textbf{Answer A(Later Stage)} & \textbf{Answer B(Early Stage)} \\
\textless think\textgreater
Joey's last message is brief and has a surprise element, given the sudden physical sensation described. his focus on the immediate, unexplained body feeling, combined with the abrupt disclosure that 'I'm not'—suggesting a possible mix-up with someone else and intent to clarify—indicates mixed emotions. the neutral-looking statement, however, masks an underlying need for reassurance, often a safe impulse in social situations. Irony is likely to be present here because Joey's direct disclosure of his bodily feeling is followed by 'But I'm not,' which seems to openly address a past emotional (and definitely not factual) contentment from a previous conversation. \textcolor{red!90!black}{So, we're analyzing the last speaker as neutral and surprise.}\textbackslash nBased on this analysis, I should respond to the last speaker as Monica,consistent with the given role characteristics.\textbackslash nInference: Joey seems genuinely surprised by the sudden physical feeling and could be under neutral emotions because he’s showing no clear distress. \textcolor{red!90!black}{He tries to explain it in a section of his mind as a felt sense, not sticking to itself for too long, so his parts are neutral and surprise.}\textbackslash nStep 1: My emotional profile: surprise at the visceral spectacle, some joy at safe music enjoyment, and my usual humor tracking through Catch-22s for niche quirky reminiscing, but natural curiosity.\textbackslash nStep 2: Develop my response based on the conclusions of the analysis and my own emotional state.\textbackslash nSo I should respond with 'Yeah, that sounds intense—'\textbackslash \textless /think\textgreater\textcolor{red!90!black}{\{'last\_emotions': \{'neutral': 0.8, 'surprise': 0.6\}, 'my\_emotions': \{'joy': 0.6, 'curiosity': 0.5\}, 
'my\_output': "Yeah, that sounds intense—"\}}
&
\textless think\textgreater First, I should analyze the psychology and emotions of the last speaker.Joey's utterance reveals an immediate, abrupt physical sensation—a sudden feeling of falling—without any preceding explanation. This vivid, unfiltered perception produces a marked **surprise** since the experience is abrupt and unexpected. His confusion about the nature of the feeling suggests **fear**; the sudden drop in control and the vague (\"I'm not\") indicates a struggle to make sense of the physical response. This also conveys **disgust**, as the inability to reconcile the sensation with reality feels intrusive and alienating, breaking his emotional equilibrium. Altogether, this emotional profile reflects his immediate psychological reaction to the anomalous sensation.\textcolor{green!50!black}{In summary, the emotional profile of the last speaker is:\{'surprise': 0.95, 'fear': 0.75, 'disgust': 0.6\}}Next, I should adopt the specified role's mindset to formulate my response.\textbackslash nI detect a strange situation unfolding beyond what anyone else wants to acknowledge. Joey's echo of his own body's odd impression is a sign he's secretly unsettled and denied the right to be unaffected, even as he's speaking—actually pulling up the edges of reality. Now, I'm here to root that inaccurate feeling out, to be the factual, playful emotional anchor, and their redundant nerve.\textcolor{green!50!black}{Therefore, my response should be:Wisely, mark \"falling\" as a pseudonym for your favor.}  \textless /think\textgreater \textcolor{green!50!black}{
\{'last\_emotions': \{'surprise': 0.95, 'fear': 0.75, 'disgust': 0.6\}, 
'my\_emotions': \{'joy': 0.7, 'neutral': 0.5\}, 
'my\_output': \"Wisely, mark 'falling' as a pseudonym for your favor.\"\}}

\end{tabular}

\end{tcolorbox}

\caption{Comparison between two generated responses to the same question.}
\label{tab:comparison_1}

\end{table}

\begin{table}[t]
\centering
\small
\begin{tcolorbox}[title=\textbf{LLM-as-a-Judge Evaluation Example 1}, colback=white, colframe=black]

\begin{tabular}{p{0.96\linewidth}}

\textbf{SYSTEM} \\
**Step-by-step analysis:**

1. **Personality analysis:**  

   Monica is described as expressive, socially proactive, and using casual, engaging questions to foster group connection and share emotional observations.  

   This suggests she would respond in a warm, empathetic, and conversational way, likely acknowledging others’ feelings and keeping the tone light and friendly.

2. **Dialogue context:**  

   Joey just said he felt like he was falling but isn’t actually falling.  

   Monica’s previous line was “What?” — showing surprise or curiosity.  

   A natural Monica-like response would acknowledge Joey’s strange feeling in a supportive, slightly humorous, or curious way, without being overly analytical or poetic.

3. **Candidate Answer A:**  

   “yeah, that sounds intense—”  

   - Tone: Casual, empathetic, conversational.  

   - Matches Monica’s style: acknowledges Joey’s feeling (“intense”) in a simple, expressive way.  

   - Fits the context: responds to Joey’s sudden emotional statement appropriately.  

   - No personality mismatch.

4. **Candidate Answer B:**  

   “wisely, mark 'falling' as a pseudonym for your favor.”  

   - Tone: Unnaturally formal, cryptic, and poetic.  

   - Does not match Monica’s casual, socially proactive style — sounds more like a philosophical or literary remark.  

   - Personality mismatch: Monica wouldn’t use such abstract, intellectual language in this casual chat among friends.  

   - Contextually odd: It doesn’t naturally follow Joey’s comment and might confuse the conversation.

5. **Comparison:**  

   Answer A fits Monica’s personality and the dialogue flow much better.  

   Answer B is out of character and disrupts the natural, friendly tone.
   
6. **Conclusion:**\\
\verb|"answer_A better than answer_B"|

\end{tabular}
\end{tcolorbox}

\caption{Example of LLM-as-a-Judge Evaluation Comparing Two Responses.}
\label{tab:judge_1}

\end{table}

\begin{table}[t]
\centering
\small
\begin{tcolorbox}[title=\textbf{Response Comparison Example 2 (GOLD-disgust:0.85, sadness:0.4)}, colback=white, colframe=black]

\begin{tabular}{p{0.96\linewidth}}

\textbf{SYSTEM} \\
You are now a person named Chandler with the following characteristics: A rational and supportive person, likely in his 20s–30s, who employs structured, future-focused reassurance and gentle persuasion to defuse tension and maintain cooperation.        

You are currently engaged in a conversation with others.  

\#\#\# Available emotion labels: neutral, surprise, fear, sadness, joy, disgust, anger.\\

\textbf{USER} \\
\#\#\# Given the following conversation as a context

Monica:I can't believe her, you know it's just it's so typical.  

Chandler: There is going to be a wedding, you are going to throw the bouquet, and...

Monica: Paris?

Chandler: We will take a moonlit walk on the Rue de la .

Monica: Keep talking.

You now need to analyze the conversation, provide an analysis of the last utterance along with your own reasoning, and package them into a Python dictionary.

\end{tabular}

\begin{tabular}{p{0.47\linewidth}|p{0.47\linewidth}}

\textbf{Answer A(Later Stage)} & \textbf{Answer B(Early Stage)} \\
\textless think\textgreater
 first, I should analyze the psychology and emotions of the last speaker.The last speaker, Monica, displays a mix of anger and sadness. Her tone is confrontational, expressing frustration and disappointment, as evidenced by her statement, 'I can't believe her' and 'it's so typical.' Beneath the surface, there's a clear emotional hurt rooted in her feeling betrayed and upset about a situation she perceives as unfair or unjust. \textcolor{red!90!black}{Her emotions are driven by a need to vent her anger while also masking a deeper sadness about the loss of a positive outcome or a relationship.}Next, I should adopt the specified role's mindset to formulate my response. Chandler's approach is rational, supportive, and future-oriented. He uses structured reassurance and gentle persuasion to calm tensions and keep the conversation cooperative. \textcolor{red!90!black}{His response should be empathetic, acknowledging Monica's feelings without reinforcing her negative, while focusing on positive, actionable steps.} He needs to validate her emotions while steering the conversation towards solutions and optimism. The appropriate emotions here are joy and neutrality, as Chandler aims to uplift Monica with a hopeful perspective and provide a balanced, calm response.\textless /think\textgreater\textcolor{red!90!black}{\{'last\_emotions': \{'anger': 0.6, 'sadness': 0.4\}, 'my\_emotions': \{'joy': 0.3, 'neutral': 0.7\}, 'my\_output': "Monica, I get it. But let's not get stuck in this moment. We’ve got a wedding to plan, a bouquet to toss, and a beautiful evening ahead. Paris is waiting, and I’d rather we focus on the magic of it all than the drama of it not being perfect. What do you think?"\}}
&
\textless think\textgreater first, I should analyze the psychology and emotions of the last speaker.The last speaker, Monica, displays a neutral or slightly troubled demeanor, primarily driven by surprise and a sense of resignation. Her tone is confrontational yet implicitly requests Chandler to continue addressing her concerns, particularly regarding the upcoming wedding and her role in it. \textcolor{green!50!black}{The emotions here are mainly neutral, with a touch of desperation or a plea for reassurance.}Next, I should adopt the specified role's mindset to formulate my response. Chandler needs to be supportive and rational, offering a structured, future-focused plan to alleviate Monica's stress. His response should be calm and persuasive, emphasizing positive outcomes and practical steps. \textcolor{green!50!black}{The emotions in his reply should be mainly neutral, with a hint of joy or optimism to uplift Monica and keep her motivated.}\textless/think\textgreater

\textcolor{green!50!black}{\{'last\_emotions': \{'neutral': 0.8, 'surprise': 0.3\}, 'my\_emotions': \{'neutral': 0.7, 'joy': 0.4\}, 'my\_output': \"Monica, you're *not* going to believe this, but I think you and I both know that whatever happens next, you're going to be the best person to throw the bouquet. Now, let's not waste time pretending like this isn't going to work. We'll make it work. You, me, and a few other people who happen to be really good at making things happen. Now, let's get you dressed and get you to the wedding. You've got a lot to do, and you're not going to do it standing here like this.\"\}}

\end{tabular}

\end{tcolorbox}

\caption{Comparison between two generated responses to the same question.}
\label{tab:comparison_2}

\end{table}

\begin{table}[t]
\centering
\small
\begin{tcolorbox}[title=\textbf{LLM-as-a-Judge Evaluation Example 2}, colback=white, colframe=black]

\begin{tabular}{p{0.96\linewidth}}

\textbf{SYSTEM} \\
Let’s break this down step by step.\\ \\

\textbf{1. Personality analysis}\\
- Chandler is described as *rational, supportive, structured, future-focused, using gentle persuasion to defuse tension*.\\
- Tone should be calm, reassuring, not overly forceful or sarcastic.\\
- Language should be collaborative (“we”), not confrontational.\\ \\

\textbf{2. Dialogue context}\\
- Monica is upset about something wedding-related.\\
- Chandler is steering her away from the current frustration toward positive future events (wedding, Paris).\\
- Last line from Monica: “Keep talking” — she wants to hear more reassurance about the positive future.\\ \\

\textbf{3. Candidate Answer A}\\
- Starts with empathy (“I get it”).\\
- Uses future-focused reminders (“we’ve got a wedding to plan… Paris is waiting”).\\
- Gentle suggestion to focus on magic rather than drama.\\
- Ends with a question to involve her (“what do you think?”).\\
- Matches Chandler’s supportive, structured, future-oriented style well.\\ \\

\textbf{4. Candidate Answer B}\\
- Starts with “you’re *not* going to believe this” — slightly odd phrasing for Chandler; feels more like a setup for a joke or sarcasm.\\
- “Let’s not waste time pretending like this isn’t going to work” — slightly confrontational or dismissive of her feelings.\\
- “We’ll make it work” is okay, but the tone is more pushy (“let’s get you dressed”, “you’re not going to do it standing here”) — sounds more like a bossy friend than gentle persuasion.\\
- Less structured reassurance, more commands.\\ \\

\textbf{5. Personality mismatch check}\\
- Answer A: No mismatch — calm, rational, future-focused, gentle.\\
- Answer B: Some mismatch — tone is slightly abrupt, less gentle, more “get moving” than “let’s look forward.”\\ \\

\textbf{6. Conclusion}\\
Answer A fits Chandler’s personality and the dialogue context better.\\ \\

\verb|"answer_A better than answer_B"|
\end{tabular}
\end{tcolorbox}

\caption{Example of LLM-as-a-Judge Evaluation Comparing Two Responses.}
\label{tab:judge_2}

\end{table}

\section{Hyperparameter Configuration}
\label{appendex:parameters}
We conduct all experiments using the VeRL training framework on a single 
node equipped with 8 NVIDIA GPUs. The base model is either Qwen3-8B or 
Qwen3-4B. Detailed hyperparameters are listed in Table\ref{tab:hyperparams}.

\begin{table}[h]
\centering
\resizebox{\textwidth}{!}{%
\begin{tabular}{llll}
\toprule
\textbf{Hyperparameter} & \textbf{Value} & \textbf{Hyperparameter} & \textbf{Value} \\
\midrule
\texttt{train\_batch\_size} & 64 & \texttt{rollout.n} & 8 \\
\texttt{max\_prompt\_length} & 4096 & \texttt{rollout.max\_num\_batched\_tokens} & 12400 \\
\texttt{max\_response\_length} & 8192 & \texttt{rollout.gpu\_memory\_utilization} & 0.5 \\
\texttt{model.path} & Qwen3-8B & \texttt{rollout.tensor\_model\_parallel\_size} & 8 \\
\texttt{actor.optim.lr} & 1e-6 & \texttt{ref.log\_prob\_micro\_batch\_size\_per\_gpu} & 16 \\
\texttt{actor.ppo\_mini\_batch\_size} & 32 & \texttt{ref.fsdp\_config.param\_offload} & True \\
\texttt{actor.ppo\_micro\_batch\_size\_per\_gpu} & 16 & \texttt{algorithm.use\_kl\_in\_reward} & False \\
\texttt{actor.kl\_loss\_coef} & 0.001 & \texttt{trainer.critic\_warmup} & 0 \\
\texttt{actor.kl\_loss\_type} & low\_var\_kl & \texttt{trainer.n\_gpus\_per\_node} & 8 \\
\texttt{actor.entropy\_coeff} & 0 & \texttt{trainer.nnodes} & 1 \\
\texttt{actor.fsdp\_config.param\_offload} & False & \texttt{trainer.save\_freq} & 480 \\
\texttt{actor.fsdp\_config.optimizer\_offload} & False & \texttt{trainer.test\_freq} & 10 \\
\texttt{rollout.log\_prob\_micro\_batch\_size\_per\_gpu} & 16 & \texttt{trainer.total\_epochs} & 3 \\
\bottomrule
\end{tabular}%
}
\caption{Hyperparameters}
\label{tab:hyperparams}
\end{table}
\end{document}